\documentclass{article}

\usepackage[dblblindworkshop, final]{neurips_2025}

\usepackage[utf8]{inputenc} 
\usepackage[T1]{fontenc}    
\usepackage{hyperref}       
\usepackage{url}            
\usepackage{booktabs}       
\usepackage{amsfonts}       
\usepackage{nicefrac}       
\usepackage{microtype}      
\usepackage{xcolor}         
\usepackage{graphicx}
\usepackage{amsmath} 
\usepackage{float}
\usepackage{enumitem}
\usepackage[normalem]{ulem}
\usepackage[most]{tcolorbox}
\usepackage{verbatim}
\usepackage{listings}
\usepackage{threeparttable}

\newtcolorbox{promptbox}[1][]{
  colback=gray!10,
  colframe=gray!50!black,
  boxrule=0.5pt,
  arc=2pt,
  left=5pt,
  right=5pt,
  top=5pt,
  bottom=5pt,
  fontupper=\small\ttfamily,
  breakable,
  #1
}

\title{InterpDetect: Interpretable Signals for Detecting Hallucinations in
Retrieval-Augmented Generation}
\workshoptitle{Workshop on Mechanistic Interpretability}

\author{%
  Likun Tan, Kuan-Wei Huang, Joy Shi, Kevin Wu\thanks{Corresponding Author} \\
  Pegasi AI, NYC\\
  \texttt{likun,kuanwei,joy,kevin@usepegasi.com} \\
}

\begin{document}

\maketitle

\begin{abstract}
Retrieval-Augmented Generation (RAG) integrates external knowledge to mitigate hallucinations, yet models often generate outputs inconsistent with retrieved content. Accurate hallucination detection requires disentangling the contributions of external context and parametric knowledge, which prior methods typically conflate. We investigate the mechanisms underlying RAG hallucinations and find they arise when later-layer FFN modules disproportionately inject parametric knowledge into the residual stream. To address this, we explore a mechanistic detection approach based on \textit{external context scores} and \textit{parametric knowledge scores}. Using Qwen3-0.6b, we compute these scores across layers and attention heads and train regression-based classifiers to predict hallucinations. Our method is evaluated against state-of-the-art LLMs (GPT-5, GPT-4.1) and detection baselines (RAGAS, TruLens, RefChecker). Furthermore, classifiers trained on Qwen3-0.6b signals generalize to GPT-4.1-mini responses, demonstrating the potential of \texttt{proxy-model evaluation}. Our results highlight mechanistic signals as efficient, generalizable predictors for hallucination detection in RAG systems. \footnote{Our code and data are available at \url{https://github.com/pegasi-ai/InterpDetect}.}
\end{abstract}

\section{Introduction}
\label{sec:intro}

Large language models (LLMs) achieve strong performance across tasks such as question answering, summarization, and code generation~\cite{wang2024infi,szalontai2024llm,jiang2024survey}, yet they frequently generate~\emph{hallucinations}—outputs that are factually incorrect or unsupported by evidence~\cite{ji2023survey}. Retrieval-Augmented Generation (RAG) aims to mitigate hallucinations by grounding outputs in external knowledge~\cite{gao2023retrieval}, but models may still produce responses that contradict retrieved content~\cite{song2024rag,hu2024lrp4rag,sun2024redeep,tan2025fred}. Detecting these hallucinations is crucial in high-stakes domains such as healthcare, finance, and education.

Recent work has advanced RAG-specific hallucination detection along two axes. First, token- and span-level corpora and toolkits, such as RAGTruth and LettuceDetect, enable fine-grained supervision and efficient classifiers~\cite{niu2024ragtruth,kovacs2025lettucedetect}, while multilingual shared tasks (Mu-SHROOM/SemEval-2025) and pipelines (HalluSearch) foster robust, cross-lingual detection~\cite{vazquez2025mushroom,boito2025hallusearch}. Second, mechanistic and attribution-based approaches, including ReDeEP and LRP4RAG, analyze internal activations to disentangle parametric (internal) knowledge from retrieved (external) context, revealing that overactive FFN pathways and attention mis-weighting often correlate with hallucinations~\cite{sun2024redeep,sun2025aarf,liu2024lrp4rag}. Surveys further synthesize best practices and open challenges in RAG hallucination detection~\cite{wang2025ragevalsurvey,zhao2025ragsurvey}.

Building on this literature, we explore the use of \textbf{External Context Score (ECS)} and \textbf{Parametric Knowledge Score (PKS)}, computed across layers and attention heads following the ReDeep framework, as predictive features for hallucination detection in the domain of financial question and answering.  Mechanistically, ECS quantifies how much a model’s output relies on retrieved information, computed via attention weights and semantic similarity between attended context and generated tokens. High ECS indicates strong grounding in external knowledge, reducing hallucination risk.
PKS reflects contributions from feed-forward networks (FFNs) and attention pathways that inject parametric knowledge into the residual stream; over-reliance can drive hallucinations. Together, ECS and PKS decompose generation into external versus internal sources, enabling fine-grained analysis of hallucination origins in RAG models.

Using Qwen3-0.6b as the base model, we compute ECS and PKS across layers and attention heads, and use them as input features to train regression-based classifiers for hallucination detection. We further demonstrate the effectiveness of \texttt{proxy-model evaluation} by applying classifiers trained on Qwen3-0.6b to outputs from larger models, such as GPT-4-mini, achieving comparable performance with substantially lower computational resource.

Our contributions are threefold:
\begin{enumerate}[left=0pt]
\item We extend the ReDeEP framework for computing \emph{external context scores} (ECS) and \emph{parametric knowledge scores} (PKS) with a fully open-sourced implementation built on TransformerLens~\cite{nanda2022transformerlens}, enabling compatibility with any TransformerLens-supported model without modifying the underlying model library.
\item We conduct a systematic evaluation of multiple regression-based classifiers, identifying the optimal model for hallucination prediction.
\item We demonstrate that leveraging a 0.6b-parameter model as a proxy allows effective and economical computation, facilitating practical application to large-scale, production-level models.
\end{enumerate}
This work highlights the utility of mechanistic signals as reliable, low-cost indicators for hallucination detection, advancing the safe and trustworthy deployment of LLMs.

\section{Related Work}
\label{sec:related}

\paragraph{Hallucination Detection in RAG:}

Hallucination detection methods span black-box classifiers, attribution, uncertainty modeling, and fine-tuning pipelines. Token-level approaches such as LettuceDetect\cite{kovacs2025lettucedetect} achieve efficient span-level detection but generalize poorly across models. Attributional methods like LRP4RAG\cite{liu2024lrp4rag} provide interpretability via relevance propagation, yet incur high computational cost. Uncertainty-based methods (e.g., FRANQ\cite{fadeeva2025franq}) quantify response faithfulness but do not explain causal mechanisms, while fine-tuning pipelines such as RAG-HAT\cite{song2024raghat} require costly data and training.
Our work builds on ReDeEP~\cite{sun2024redeep}, the first mechanistic framework for RAG hallucinations, which disentangles parametric (internal) and contextual (retrieved) contributions through \emph{external context} and \emph{parametric knowledge} scores. We extend this line by providing a TransformerLens-based implementation, systematically benchmarking regression-based classifiers, and showing that signals extracted from a 0.6b model transfer effectively to larger models. This retains ReDeEP’s interpretability while improving scalability and practicality for deployment.

\paragraph{Parametric vs. External Knowledge: }
Recent mechanistic interpretability studies have elucidated how LLMs balance \emph{parametric knowledge} (FFN-stored internal memory) and \emph{external context} (retrieved information via attention heads). Hallucinations often arise when later-layer FFNs over-inject parametric knowledge into the residual stream while attention underweights retrieved content, causing misalignment between internal and external sources~\cite{sun2024redeep}. Models also exhibit a ``shortcut'' bias, over-relying on retrieved context even when parametric knowledge is complementary~\cite{ghosh2024reliance}. Knowledge is structured across neurons, with semantically related information clustered in FFNs and attention heads~\cite{yu2024neuron,cheng2024interplay}. While our work focuses on RAG hallucination detection, these findings support broader applications such as targeted knowledge editing and controlled grounding, enhancing interpretability and reliability in high-stakes LLM deployments~\cite{parry2025mechir}.

\section{Methodology}
\label{sec:method}

\begin{figure}[H]  
    \centering
    \includegraphics[width=1.0\textwidth]{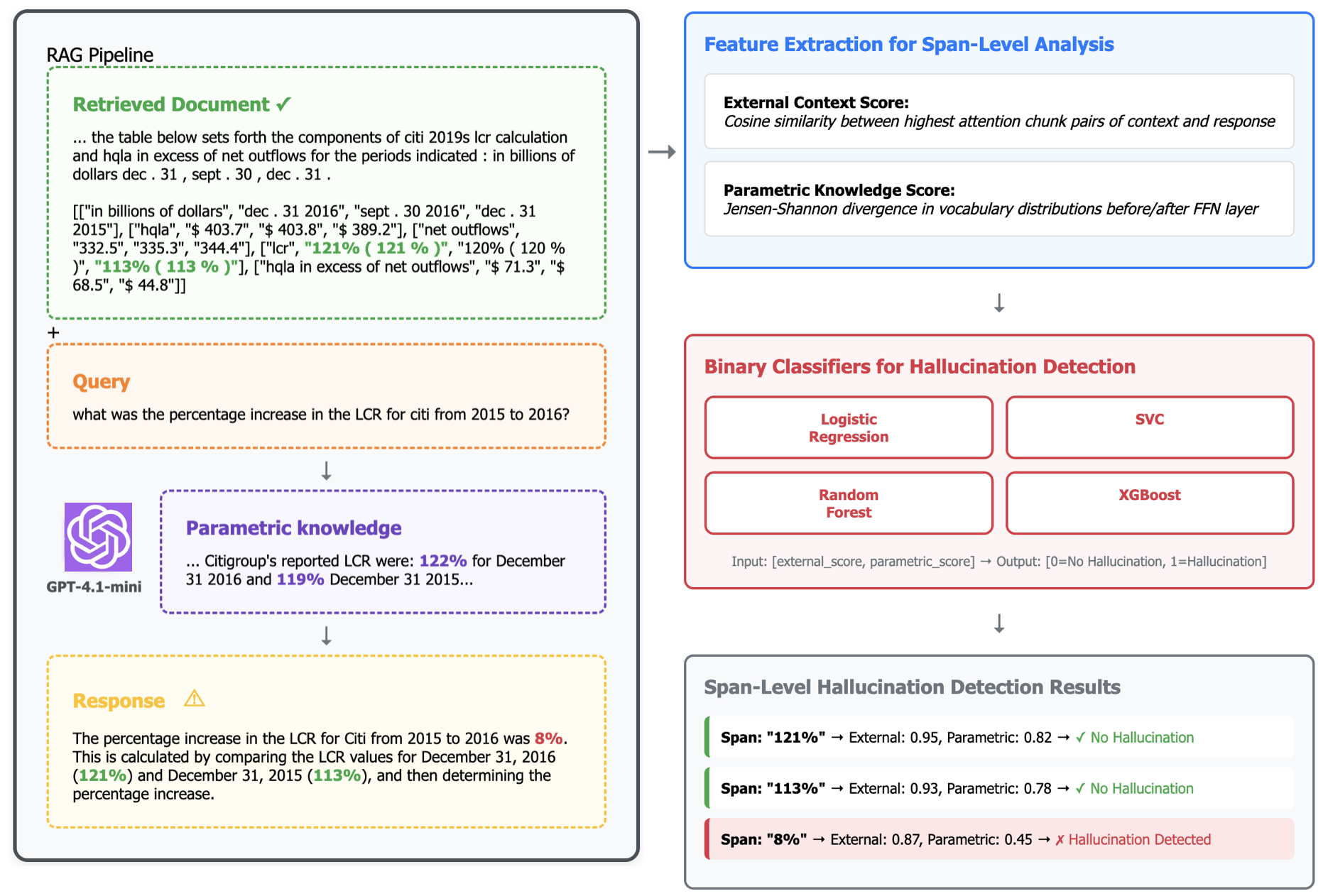}
    \caption{Pipeline for Span-Level Hallucination Detection}
    \label{fig:method}
\end{figure}

The main stages of our methodology are illustrated in Figure~\ref{fig:method}. We first follow a standard RAG pipeline to generate a response given a query and retrieved document. The model’s \emph{parametric knowledge} is treated as internal knowledge, elicited by prompting the LLM with the query. For each context-response span pair, we compute the \emph{External Context Score} (ECS) for attention heads and the \emph{Parametric Knowledge Score} (PKS) for feed-forward network layers, capturing contributions from external and internal knowledge, respectively. After confirming alignment with hallucination labels via correlation analysis, these scores are used as features for binary classifiers to detect hallucinated spans, which are then aggregated to yield response-level predictions. The following sections provide a more detailed discussion of each step.

\subsection{Data Curation}
\label{subsec:data}

We utilize the FinQA dataset from RagBench~\cite{friel2024ragbench} as our primary source. The dataset comprises 12.5k training instances and 2.29k test instances. Each instance includes a document drawn from the financial reports of S\&P 500 companies, a question whose answer is grounded in the document, and a response originally generated by gpt-3.5-turbo and claude-3-haiku. For the purposes of our study, we regenerate responses using Qwen3-0.6b---the model consistently employed for computing interpretability scores---as well as GPT-4.1 mini, in order to examine whether the proposed method extends to hallucination detection in outputs from alternative models, thereby reflecting a more practical setting. Our data generation is divided into three steps as follows:

\begin{itemize}[left=0pt]
\item \textbf{Response Generation:} We subsample 3{,}000 training instances from the original dataset. Responses are generated using Qwen3-0.6b and GPT-4.1 mini, with each model producing outputs separately. 
\item \textbf{Labeling:} To train the hallucination detection model, we need to identify spans in responses that contain hallucinated content relative to the source document. We first use LettuceDetect~\cite{kovacs2025lettucedetect} for span-level labeling. To address potential errors, we add two LLM-based judges: llama-4-maverick-17b-128e-instruct and gpt-oss-120b, which evaluate responses generated by Qwen3-0.6b. For responses from the GPT family, we replace gpt-oss-120b with claude-sonnet-4 to reduce family-specific bias. We then apply majority voting at the response level. A sample is kept if at least one additional judge agrees with the LettuceDetect label. After filtering, 1{,}852 samples remain for downstream analysis. 
\item \textbf{Chunking:} To improve the accuracy of external context and parametric knowledge scores, we compute them at the span level. Therefore, we chunk both the retrieved document and the response. For the retrieved document, we use the existing \texttt{documents\_sentences} from the original dataset. For responses, we split them into individual sentences.
\end{itemize}
An example from data preprocessing is provided in Appendix~\ref{appx:ex}.

\subsection{Mechanistic Metrics}
\label{subsec:metrics}

Our work draws on concepts from mechanistic interpretability research. Specifically, we leverage TransformerLens, an open-source library that provides access to internal model parameters—such as attention heads, feed-forward network layers (FFNs), and residual streams—in GPT-like models. In these models, residual connections allow each layer to incrementally update the hidden state using information from attention heads and FFNs. For a detailed description of the architecture, we refer the reader to the original TransformerLens work~\cite{nanda2022transformerlens}.
We primarily use TransformerLens to compute two metrics: the \emph{External Context Score} and the \emph{Parametric Knowledge Score}, following the definitions in ReDeEP~\cite{sun2024redeep}. While the original ReDeEP framework supports both token-level and chunk-level hallucination detection, computing scores at the token level is computationally expensive and does not fully capture context. Therefore, we restrict our calculations to the chunk level for both metrics.

\paragraph{External Context Score:} The External Context Score (ECS) quantifies the extent to which a language model leverages external context when generating a response. In mechanistic interpretability, attention heads are responsible for retrieving relevant information from the context. To measure this utilization, ECS captures the semantic alignment between response segments and the context chunks most strongly attended to by the model.

Let the external context be partitioned into chunks $\{\tilde{c}_1, \tilde{c}_2, \dots, \tilde{c}_M\}$, and the generated output into response chunks $\{\tilde{r}_1, \tilde{r}_2, \dots, \tilde{r}_N\}$, corresponding to \texttt{prompt\_spans} and \texttt{response\_spans} in Appendix~\ref{appx:ex}. For each attention head $h$ at layer $l$, the most relevant context chunk for each response chunk $\tilde{r}_j$ is identified as
\begin{align*}
\tilde{c}_j^{\,\ell,h} &= \arg\max_{\tilde{c}_i} A(\tilde{r}_j^{\,\ell,h}, \tilde{c}_i^{\,\ell,h}),
\end{align*}
where $A$ denotes the token-level attention weight matrix, and $l$ and $h$ indicate the layer and head indices. The chunk-level ECS is defined as the cosine similarity between the embeddings of $\tilde{r}_j$ and its corresponding context chunk $\tilde{c}_j$:
\begin{align*}
\text{ECS}_{\tilde{r}_j}^{\,\ell,h} &= \cos\big( e(\tilde{r}_j^{\,\ell,h}), \, e(\tilde{c}_j^{\,\ell,h}) \big),
\end{align*}
where $e(\cdot)$ represents the embedding function, and $\cos(\cdot, \cdot)$ denotes cosine similarity. 

\paragraph{Parametric Knowledge Score:} The Parametric Knowledge Score (PKS) quantifies the extent to which the FFN contributes to \emph{parametric knowledge}, i.e., knowledge stored in the model's weights, as opposed to information coming from external context. This is done by measuring the difference between residual stream states \textbf{before} the FFN layer and \textbf{after} the FFN layer.  

Since the residual stream itself does not directly indicate "which token is being suggested"—it is a latent vector—we map it through the unembedding/projection matrix to the vocabulary distribution. This allows us to observe how the residual change after the FFN influences predicted token probabilities.  

We then apply the Jensen-Shannon divergence (JSD) to compute the distance between the two vocabulary distributions, which defines the token-level PKS. The chunk-level PKS is computed by averaging the token-level PKS over all tokens in the chunk. Mathematically, this is expressed as
\begin{align*}
\text{PKS}_{t_n}^\ell = \operatorname{JSD}\big( p(x_n^{\text{mid},\ell}), \, p(x_n^\ell) \big),  \hspace{5pt}  \text{PKS}_{\tilde{r}}^\ell = \frac{1}{|\tilde{r}|} \sum_{t_n \in \tilde{r}} \text{PKS}_{t_n}^\ell.
\end{align*}
where $p(\cdot)$ denotes the mapping from residual stream states to vocabulary distributions, and $x_n^{\text{mid},\ell}$ and $x_n^\ell$ refer to the residual stream states before and after the FFN layer, respectively. $\text{PKS}_{t_n}^\ell$ and $\text{PKS}_{\tilde{r}}^\ell$ stand for token-level and chunk-level PKS, respectively.

More details about the computation is given in Appendix~\ref{appx:metric}.

\subsection{Classifier for Hallucination Detection}
\label{subsec:clf}

Hallucination detection is formulated as a binary classification task. As input features, we use the \textbf{External Context Score} (ECS) computed for each attention head and layer, together with the \textbf{Parametric Knowledge Score} (PKS) computed for each layer. Prior to classification, features are standardized using \texttt{StandardScaler} and refined via \texttt{SmartCorrelatedSelection} to remove redundant or highly correlated features. We evaluate four classifiers—Logistic Regression, Support Vector Classification (SVC), Random Forest, and XGBoost—and select the model achieving the best performance for inference. Predictions are generated at the span level and can subsequently be aggregated to obtain response-level hallucination detection results.

\section{Experiments}\label{sec:experiments}

\subsection{Correlation Analysis}\label{subsec:correlation}

The primary objective of this work is to leverage mechanistic signals, i.e., the \textbf{External Context Score} (ECS) and the \textbf{Parametric Knowledge Score} (PKS), for hallucination detection, under the assumption that both are correlated with hallucination occurrence in generated responses.  

\begin{figure}[htbp]  
    \centering
    \includegraphics[width=0.9\textwidth]{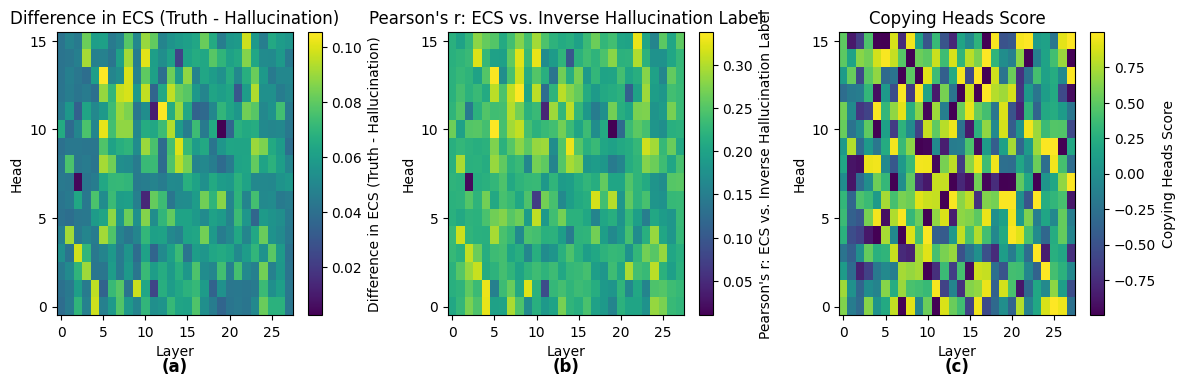}
    \caption{Relationship Between LLM Utilization of External Context and Hallucination}
    \label{fig:ECS}
\end{figure}

We begin by examining the relationship between ECS and RAG hallucinations. Specifically, we compare ECS values between truthful and hallucinated responses. Figure~\ref{fig:ECS}(a) reports per-layer, per-head scores, all of which are positive. Since ECS reflects an LLM’s reliance on retrieved context through attention heads, these results indicate that hallucinated responses utilize less external context than truthful ones. To further test this hypothesis, we computed the Pearson Correlation Coefficient (PCC) between hallucination labels and ECS. Because a negative correlation was expected, we use the inverse hallucination label instead. As shown in Figure~\ref{fig:ECS}(b), all attention heads exhibit negative correlations, confirming that higher ECS values are associated with lower hallucination likelihood. Taken together, Figures~\ref{fig:ECS}(a) and \ref{fig:ECS}(b) suggest that RAG hallucinations emerge when the model fails to adequately exploit external context.  We also examine the role of copying heads by computing \texttt{OV\_copying\_score} as a proxy of \texttt{full\_OV\_copying\_score} (Figure~\ref{fig:ECS}(c)). This proxy is based on the findings in~\cite{transformerlens_scoring_2025}, which shows that \texttt{OV\_copying\_score} and \texttt{full\_OV\_copying\_score} exhibit a strong positive correlation across heads and layers, suggesting that \texttt{OV\_copying\_score} can provide interpretable insights at lower computational cost. However, unlike the strong correlation observed between ECS and copying head scores in LLaMA models~\cite{sun2024redeep}, we did not find such correlation in Qwen3-0.6b. Consequently, we use attention scores from all layers and heads for ECS calculation and rely on feature-selection techniques during the classification stage.  

\begin{figure}[htbp]  
    \centering
    \includegraphics[width=0.9\textwidth]{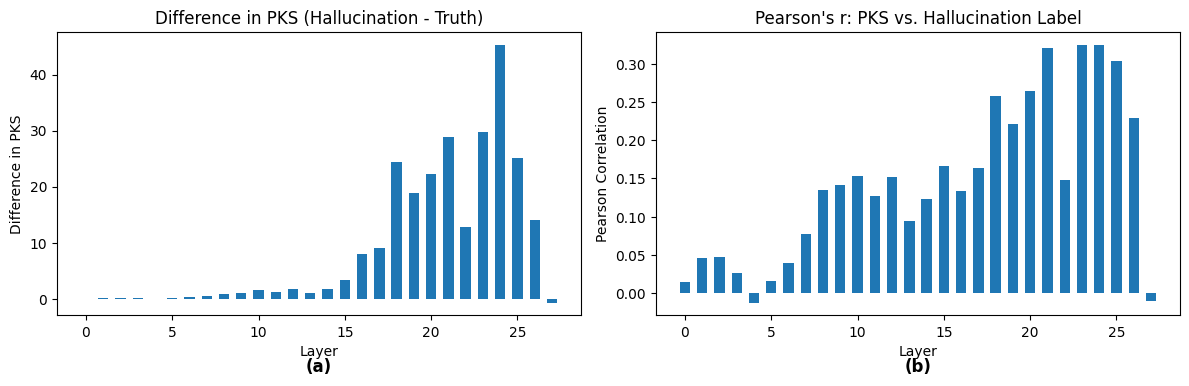}
    \caption{Relationship Between LLM Utilization of Parametric Knowledge and Hallucination}
    \label{fig:PKS}
\end{figure}

We next investigate how PKS contributes to RAG hallucinations. Figure~\ref{fig:PKS}(a) shows that PKS values are positive across nearly all layers, except the final one. Notably, FFN modules in later layers exhibit substantially higher scores for hallucinated responses than for truthful ones, yielding elevated layer-averaged scores for hallucinations. Consistently, Figure~\ref{fig:PKS}(b) presents the Pearson correlation between hallucination labels and PKS, revealing that later-layer FFNs are positively correlated with hallucinations.

\subsection{Hallucination Detection} \label{subsec:detect}

\paragraph{Training of Classifiers:} Training was conducted at the span level. Following the data preparation procedure in Section~\ref{subsec:data}, we obtained 1{,}852 instances, corresponding to 7{,}799 span-level samples (4{,}406 negative and 3{,}393 positive). The input features consist of the \textbf{External Context Score} (ECS) and \textbf{Parametric Knowledge Score} (PKS) defined in Section~\ref{subsec:metrics}. For Qwen3-0.6b, which has 28 layers and 16 attention heads per layer, this yields 476 features in total. To mitigate redundancy, feature selection reduced the dimensionality from 476 to 341. Results are reported in Table~\ref{tab:clf}. Among the four models, SVC achieved the highest validation F1 score and was selected as the final prediction model. By contrast, XGBoost achieved strong training performance but exhibited severe overfitting, highlighting the risk of high-capacity models with limited data and the need for stronger regularization or substantially more training samples to generalize effectively. In comparison, SVC offered a more stable balance between model complexity and generalization. Further training details are provided in Appendix~\ref{appx:training}.

\paragraph{Prediction:}  We obtain response-level labels using the trained SVC model by aggregating span-level predictions: a response is labeled as hallucinated if any span is predicted as such. We evaluate two settings: \textbf{self-evaluation}, where responses are generated by the same model used to compute ECS/PKS (Qwen3-0.6b), and \textbf{proxy-based evaluation}, where responses are produced by a different, typically production-level model (e.g., GPT-4.1 mini). We assume that, for responses grounded in retrieved context, the model should rely more heavily on external context than on its parametric knowledge. This assumption is model-agnostic. Consequently, given a pair consisting of a response and its corresponding retrieved context, we can leverage internal signals from any model to assess the relative utilization of external context versus parametric knowledge. This assumption forms the foundation of our \textbf{proxy-based} approach.

Our method is compared against a diverse set of baselines, including proprietary models (GPT-5, GPT-4.1), open-source models (GPT-OSS-20b, LLaMA-3.3-70b-versatile, LLaMA-3.1-8b-instant, Qwen3-32b, Qwen3-0.6b), and commercial detection systems (RAGAS~\cite{es2024ragas}, TruLens~\cite{datta22a}, RefChecker~\cite{hu2024refchecker}). For language models, we apply the prompt in Appendix~\ref{appx:prompt} to assess the faithfulness of responses relative to retrieved documents. For commercial detection systems, implementation details are provided in Appendix~\ref{appx:baselines}. Detection results are summarized in Table~\ref{tab:detection}, and key observations are discussed below:

\begin{itemize}[left=0pt]
\item Overall, our method achieves moderate performance. In the self-evaluation setting, we obtain higher F1 scores than TruLens and llama-3.1-8b-instant, and perform comparably to RefChecker. In the proxy-based evaluation, our method performs even better, outperforming nearly all models except GPT-5 and RAGAS in F1 score. Notably, the models with superior performance are either proprietary or significantly larger, whereas we employ an efficient classifier leveraging signals from a 0.6b parameter model.
\item We also include Qwen3-0.6b as a detection model to demonstrate that, by itself, the 0.6b model is insufficient for hallucination detection. However, when combined with a strong classifier that leverages its internal signals, performance is substantially improved.
\item Our model exhibits higher recall than precision in both settings. One possible explanation is that the training dataset may contain some false-positive samples, which could bias the classifier toward predicting more hallucinations. During data curation (see Section~\ref{subsec:data}), we retained instances where LettuceDetect won the majority vote but did not exclude low-confidence span-level labels. These low-confidence spans are potential false positives but warrant further investigation. Nonetheless, we argue that higher recall is generally desirable, as false-positive cases can be further verified in downstream tasks.
\item In proxy-based evaluation, we observe that all language models (from tiny 0.6b to GPT-5) exhibit higher precision than recall. This trend may stem from models such as GPT-4.1-mini producing fluent, reasonable-sounding responses, where subtle inaccuracies are less likely to be flagged by the detection models. In contrast, during self-evaluation, errors or unsupported content generated by Qwen3-0.6b are more readily detected by stronger models.
\end{itemize}

\begin{table}[htbp]
\centering
\caption{Span-level Detection performance (\%)}
\label{tab:clf}
\begin{tabular}{lrrrrrr}
\toprule
\textbf{Classifier} & \textbf{Train Prec.} & \textbf{Val Prec.} & \textbf{Train Rec.} & \textbf{Val Rec.} & \textbf{Train F1} & \textbf{Val F1} \\
\midrule
LR & 79.71 & 74.84 & 77.05 & 71.09 & 78.36 & 72.92 \\
SVC & 84.44 & 79.00 & 79.24 & 74.34 & 81.76 & 76.60 \\
RandomForest & 79.87 & 74.92 & 76.13 & 72.27 & 77.95 & 73.57 \\
XGBoost & 99.74 & 76.45 & 99.77 & 73.75 & 99.75 & 75.08 \\
\bottomrule
\end{tabular}
\end{table}

\begin{table}[htbp]
\centering
\begin{threeparttable}
\caption{Response-level Detection Performance (\%)}
\label{tab:detection}
\begin{tabular}{lcccccc}
\toprule
 & 
\multicolumn{3}{c}{\textbf{Self-Evaluation}} & 
\multicolumn{3}{c}{\textbf{Proxy-based Evaluation}} \\
\cline{2-7}
\textbf{Model} & \textbf{Precision} & \textbf{Recall} & \textbf{F1} & \textbf{Precision} & \textbf{Recall} & \textbf{F1} \\
\midrule
GPT-5     & 77.27 &  \textbf{92.97} &  \textbf{84.40} & 91.67 & 66.27 & \textbf{76.92}\\
GPT-4.1 &  76.39 &  85.94 &  \underline{80.88} & \textbf{94.29} & 39.76 & 55.93 \\
GPT-OSS-20b & 82.79 & 78.91 & 80.80 & 92.31 & 43.37 & 59.02\\
llama-3.3-70b-versatile & 81.03 & 73.44 & 77.05 & \underline{93.75} & 18.07 & 30.30\\
llama-3.1-8b-instant & 69.23 & 49.22 & 57.53 & 70.37 & 22.89 & 34.55 \\
Qwen3-32b & 79.55 & 82.03 & 80.77 & 86.11 & 37.35 & 52.10 \\
Qwen3-0.6b & 70.27 & 20.31 & 31.52 & 78.79 & 31.33 & 44.83\\
RAGAS & 68.45  & \underline{89.84} & 77.70 & 75.29 & 77.11 & \underline{76.19}\\
TruLens & \textbf{89.61} & 53.91 & 67.32 & 49.08 & \textbf{96.39} & 65.04\\
RefChecker & \underline{84.62} & 68.75 & 75.86 & 71.43 & 12.05 & 20.62\\
Ours & 63.89 & \underline{89.84} & 74.68 & 62.90 & \underline{93.98} & 75.36 \\
\bottomrule
\end{tabular}
\begin{tablenotes}
\footnotesize
\item \textbf{Note:} RAGAS, TruLens and RefChecker use GPT-4.1 under the hood. In all the experiments, we obtain mechanistic metrics from Qwen3-0.6b. Model response is from Qwen3-0.6b under Self-Evaluation while from GPT-4.1 mini under Proxy-based Evaluation.
\end{tablenotes}
\end{threeparttable}
\end{table}

\section{Conclusion}

In this work, we developed a detection method for RAG hallucinations by decoupling the attributions from parametric knowledge and external context. Our correlation study shows that hallucinations arise from insufficient utilization of external context and over-reliance on parametric knowledge. Guided by these insights, we experimented with a number of classification methods to predict span-level hallucination and aggregate the results for response-level detection. We demonstrate the comparable ability of this cost-free, low-memory model with the commercial counterparts. Moreover, we show that our model can be used as a proxy on evaluation of large-scale, production-level models.

\section{Limitations}\label{sec:limit}

While our approach demonstrates the effectiveness of mechanistic signals for hallucination detection in RAG, several limitations remain. First, our ECS and PKS computations rely on information from all layers and heads, which is computationally intensive, particularly during the projection from hidden states to the vocabulary distribution. Given the weak and unclear correlation between copying-head scores in Qwen3-0.6b and hallucination, future work should aim to identify the layers most critical for efficient hallucination detection. For instance, our correlation analysis suggests that retaining only late-layer signals may suffice for downstream classification, reducing computation while preserving predictive power.
Second, we only utilize ECS and PKS as input features for classification. Although these mechanistic features provide valuable signals, the overall performance remains below that of large-scale models. Expanding the feature set to include uncertainty-based measures and representation-level features could enrich the representation and improve detection accuracy.
Finally, a more ambitious application of mechanistic signals is in response intervention, where the goal is to produce more reliable and truthful outputs by steering the model during generation. In such a setting, the current \texttt{proxy-based evaluation} framework is insufficient, as effective intervention requires access to and manipulation of layer-level activations in real time. Implementing such interventions would necessitate substantial computational resources and more sophisticated tooling to interface with the model’s internal states.



\begin{ack}
We thank Neo for supporting this research through their startup accelerator program. Their contribution played a crucial role in enabling the development and evaluation of our models.

\end{ack}

\bibliographystyle{unsrt}  
\bibliography{main} 


\appendix

\section{Preprocessing Example}\label{appx:ex}

A sample entry from our preprocessed dataset is shown below:

\begin{lstlisting}[language=Python]
question: "What is the growth rate in CCG operating income in 2015?"

documents: [
  "Management 2019's discussion and analysis of financial condition and results of operations (continued). "
  "The following results drove changes in CCG operating income by approximately the amounts indicated:"
]

prompt: "Given the context, please answer the question based on the provided information from the context. Include any reasoning with the answer.\n\nContext: Management 2019's discussion and analysis of financial condition and results of operations (continued). The following results drove changes in CCG operating income by approximately the amounts indicated:\nQuestion: What is the growth rate in CCG operating income in 2015?\nAnswer:"

response: "The exact growth rate is derived from the differences in the values provided. Thus, the answer is that the growth rate in CCG operating income in 2015 is approximately -20.7%."

labels: [
  {start: 673, end: 722, confidence: 0.8037, text: "Growth Rate} = \\frac{-2161}{10327} \\times 100\\%"},
  {start: 730, end: 734, confidence: 0.5833, text: "20.7"}
]

labels_llama: 1
labels_gpt: 1

prompt_spans: [[0,135],[136,144],[145,348],[349,1177],[1178,3465]]
response_spans: [[0,383],[384,597],[598,663]]
\end{lstlisting}

\textbf{Notes:}
\begin{itemize}[left=0pt]
    \item \texttt{labels} provide span-level hallucination annotations with start and end indices in the response text.
    \item \texttt{labels\_llama} and \texttt{labels\_gpt} indicate whether the respective LLM judges marked the response as hallucinated (1) or not (0).
    \item \texttt{prompt\_spans} and \texttt{response\_spans} segment the prompt and response into sentence or phrase-level chunks for span-level scoring.
\end{itemize}

\section{Computation Details of Mechanistic Metrics}\label{appx:metric}

The computation was performed using the TransformerLens library on the Qwen3-0.6B model. Inference was executed on a Google Colab L4 GPU with 24-GB memory, using \texttt{torch.float16} precision to reduce activation storage costs. Sentence-level semantic similarity was computed using the BAAI/bge-base-en-v1.5 encoder, also hosted on GPU to avoid transfer overhead. 

The average execution time for an end-to-end ECS/PKS computation was 42 seconds per example. GPU allocation remained within 1.9–2.1 GB, with reserved memory peaking at 2.2 GB across iterations. After each iteration, tensors were explicitly released, and memory was reclaimed using \texttt{torch.cuda.empty\_cache()} and \texttt{torch.cuda.ipc\_collect()} to prevent fragmentation.

\section{Training Details of Classifiers}\label{appx:training}

We split the dataset of 1{,}852 instances (7{,}799 span-level samples in total, with 4{,}406 negative and 3{,}393 positive labels) into training and validation subsets using a 90/10 stratified split based on the hallucination label. As features, we combined external context scores (per attention head and layer) with parametric knowledge scores (per layer). Feature preprocessing was performed through a pipeline consisting of standardization (\texttt{StandardScaler}), removal of near-constant features (\texttt{DropConstantFeatures}), elimination of duplicate features (\texttt{DropDuplicateFeatures}), and correlation-based selection (\texttt{SmartCorrelatedSelection}) with a Pearson correlation threshold of 0.9. The correlation filter employed a \texttt{RandomForestClassifier} with maximum depth 5 as the estimator. After preprocessing, the feature dimensionality was reduced from the full set to a filtered subset.
We then evaluated four classifiers—Logistic Regression, Support Vector Classification (SVC), Random Forest, and XGBoost—using pipelines composed of the preprocessing module followed by the respective classifier. For Random Forest and XGBoost, we set the maximum tree depth to 5 while leaving other hyperparameters at their default values.

\section{Prompt for Baselines}\label{appx:prompt}

Below is the prompt used for hallucination detection when using models GPT-5, GPT-4.1, GPT-OSS-20b, LLaMA-3.3-70b-versatile, LLaMA-3.1-8b-instant, Qwen3-32b and Qwen3-0.6b.

\begin{promptbox}[title=User Prompt]
You are an expert fact-checker. Given a context, a question, and a response, your task is to determine if the response is faithful to the context.

    Context: 
    {\it{context}}
    
    Question:
    {\it{question}}
    
    Response:
    {\it{response}}
    
Is the response supported and grounded in the context above? Answer "Yes" or "No", and provide a short reason if the answer is "No". Be concise and objective.
\end{promptbox}

\section{Implementation of Commercial Detection Systems}\label{appx:baselines}

We describe our implementation of three commercial tools—RAGAS, TruLens, and RefChecker—for hallucination detection below.

\paragraph{RAGAS:} We use RAGAS's \texttt{faithfulness} metric to evaluate how well a model's responses align with the provided context documents. GPT-4.1 is used as the evaluator model. For each data point, a faithfulness score between 0 and 1 is computed. To determine the optimal threshold that maximizes F1 score, we evaluate F1 across candidate thresholds in [0,1]. Predictions are binarized at each threshold, and the threshold that maximizes F1—balancing precision and recall—is selected.

\paragraph{TruLens:} TruLens provides a framework for evaluating hallucination via its groundedness feedback mechanism. We use the \texttt{groundedness\_measure\_with\_cot\_reasons} function to compute groundedness scores in [0,1]. The F1-optimal threshold is determined using the same procedure as in RAGAS.

\paragraph{RefChecker:} RefChecker extracts factual claims from model responses and verifies them against reference documents. It consists of two components: an \texttt{LLMExtractor} that extracts claims, and an \texttt{LLMChecker} that classifies claims as \texttt{Entailment}, \texttt{Neutral}, or \texttt{Contradictory}. A response is labeled hallucinated if any claim is \texttt{Contradictory}, and non-hallucinated if all claims are either \texttt{Entailment} or \texttt{Neutral}.

\end{document}